\documentclass[conference]{IEEEtran}

\usepackage{graphicx}       

\usepackage{extarrows}

\title{Is the Lecture Engaging? Lecture Sentiment Analysis for Knowledge Graph-Supported 
Intelligent Lecturing Assistant (ILA) System
}

\author{\IEEEauthorblockN{Yuan An, Samarth Kolanupaka\textsuperscript{\textsection}, 
Jacob An\textsuperscript{\textsection}, Matthew Ma\textsuperscript{\textsection}, Unnat Chhatwal\textsuperscript{\textsection}, 
Alex Kalinowski,  Michelle Rogers}
\IEEEauthorblockA{\textit{College of Computing and Informatics} \\
\textit{Drexel University}\\
Philadelphia, PA 19104, USA \\
\{ya45, ajk437, mlr92\}@drexel.edu}
\and 
\IEEEauthorblockN{Brian Smith}
\IEEEauthorblockA{\centerline{\textit{Lynch School of Education and Human Development}} \\
\textit{Boston College}\\
Boston, MA 02467, USA\\
b.smith@bc.edu}
}

\begin{document}
\maketitle

\begingroup\renewcommand\thefootnote{\textsection}
\footnotetext{Drexel summer interns contributing equally to this project.}
\endgroup

\begin{abstract}
This paper introduces an intelligent lecturing assistant (ILA) system that utilizes a knowledge graph to represent course content and optimal 
pedagogical strategies. The system is designed to support instructors in enhancing student learning through real-time 
analysis of voice, content, and teaching methods. As an initial investigation, we present a case study on lecture voice sentiment analysis, 
in which we developed a training set comprising over 3,000 1-minute lecture voice clips. 
Each clip was manually labeled as either 
engaging or non-engaging. Utilizing this dataset, we constructed and evaluated several classification models based on a variety of 
features extracted from the voice clips. The results demonstrate promising performance, achieving an F1-score of 90\% for boring lectures
on an independent set of over 800 test voice clips. This case study lays the groundwork for the development of a 
more sophisticated model that will integrate content analysis
and pedagogical practices. Our ultimate goal is to aid instructors in teaching more engagingly and effectively by leveraging modern 
artificial intelligence and big data techniques.
\end{abstract}

%
%
\section{Introduction}
\label{sec:introduction}

Engaging students and facilitating the retention of knowledge in long-term memory are fundamental aspects 
of effective teaching, yet many lectures often fail to achieve this goal. Although extensive research in cognitive 
science \cite{powerful-teaching} and neuroscience \cite{oakley-uncommon} has suggested a multitude of 
scientifically-based strategies for effective teaching, the application of these findings in real classrooms remains 
limited. Advancements in artificial intelligence and big data offer a promising 
avenue for transforming the science of learning to real-world experience. This paper introduces a novel knowledge 
graph-supported intelligent lecturing assistant (ILA) system designed to help teachers enhance student learning during 
lectures by integrating insights from scientifically-based teaching strageties. 
By leveraging the power of AI, this system aims to empower 
instructors and create more interactive and engaging learning environments, ultimately 
contributing to improved student learning outcomes.
 
Agarwal and Bain in \cite{powerful-teaching}
describe four powerful scientifically based teaching strategies \cite{powerful-teaching}: \emph{Retrieval practice, spaced practice, interleaving}, 
and \emph{feedback-driven metacognition}. These strategies are drawn on empirical research by cognitive scientists and practical strategies from 
educators around the world. 
\emph{Retrieval practice} involves actively recalling information from memory, rather than simply reviewing or re-reading material. 
\emph{Spaced practice} involves spreading study sessions over time rather than cramming all at once. 
\emph{Interleaving} is the practice of mixing different topics or types of problems within a single study session, rather than focusing on one subject or skill at a time. 
\emph{Feedback-driven metacognition} involves using feedback to guide students in reflecting on their learning processes and outcomes. 

Additionally, Oakley et al in \cite{oakley-uncommon} introduce 
\emph{“Learn it, link it”}, another fundamental concept of learning  grounded in neuroscience research.
The concept explains the establishment, reinforcement, and expansion of neural connections in the neocortex of 
the learner’s brain. In educational contexts, learning involves two critical 
types of memory: \emph{working memory}, which temporarily holds information, and \emph{long-term 
memory}, where information can potentially persist for a lifetime. A primary goal of 
teaching is to facilitate the transfer of content from working memory to long-term memory. 
Given the limited capacity of working memory, effective teaching method requires breaking down 
and reviewing content at optimally short intervals to ensure its successful integration 
and retention in long-term memory. 

Each of these scientifically grounded strategies and methods involves deliberate practice and skills. 
Specifically, a teacher needs to design low-stakes quizzes, formulate content-relevant practice tests, establish study schedules that 
revisit material at progressively increasing intervals, split learning into shorter and more frequent sessions, and offer timely 
and specific feedback on quizzes and assignments.
Developing an intelligent lecturing assistant system to support teachers in 
applying these teaching strategies is both practical and beneficial.
However, for such a system to be effective, it must be capable of analyzing 
lecture sentiment and determining the optimal moments for retrieval practice. 
Additionally, the system must possess comprehensive knowledge of the entire curriculum to 
effectively apply spacing and interleaving techniques.

In this paper, we detail the development of an approach that can classify lecture voice sentiment  
for an intelligent lecturing assistant (ILA)
system which leverages a knowledge graph to represent the curriculum and
learning objectives. With the insights from the lecture sentiment analysis, the ILA system will be able to assist
teachers with timely suggestions for pauses 
and relevant quizzes and questions to reinforce student learning and memory transfer. 
Our main contributions in this study include:

\begin{itemize}
\item We created a unique training data set comprising over 3,000 1-minute lecture voice clips, 
and a unique validation set of over 800 independent voice clips. 
Each clip was manually labeled as either engaging or non-engaging for sentiment analysis.
\item We extracted a range of features from the voice clips, including temporal characteristics, perceptual
representations, spectral features, and pitch and harmonic content. 
\item Utilizing this dataset and features, we constructed and evaluated a variety of classification models. 
The results demonstrate promising performance, achieving an F1-score of 90\% 
for boring lectures on the independent validation set.
\item This case study lays the groundwork for the development of a more sophisticated model that 
will integrate content analysis and pedagogical practices. 
Our ultimate goal is to aid instructors in teaching more engagingly and effectively by leveraging modern 
artificial intelligence and big data techniques.
\end{itemize}

The rest of the paper is organized as follows. 
Section \ref{sec:related_work}  discusses related work. 
Section \ref{sec:background}  presents a background for the intelligent lecturing system. 
Section \ref{sec:problem_sentiment_analysis} formalizes the problem of lecture voice sentiment analysis.
Section \ref{sec:data_collection} details the process of collecting training and independent validation data.
Section \ref{sec:feature_extraction} presents a range of features extracted from raw acoustic signals for down-stream classification. 
Section \ref{sec:model_building} describes various classification models. 
Section \ref{sec:evaluation} presents the evaluation results. 
Section \ref{sec:discussion} discusses the findings and makes suggestions. 
Section \ref{sec:conclusion} concludes the paper. 
Finally, the Appendix presents the link to the public Github repository containing the source code and
the method for downloading the training and validation data sets.

%
%
\section{Related Work}
\label{sec:related_work}

There is a long history of developing Intelligent educational systems (IES). 
Cumming and Self's seminal paper in 1990 conceptualized IES as engaging learners 
through multiple levels of conversational interaction \cite{Cumming1990}.
Since then, a substantial body of literature has emerged, addressing various aspects and technologies of IES 
 \cite{towards-design-IES,AI-ITS-sustainable-education,multisensory-interaction-review,Chaudhry2021-AIEd,Turakhia2024-understanding}. 
 The primary focus of IES development has been on intelligent tutoring systems (ITS)
 \cite{ITS-historical-survey,Wang2023-ITS}, with recent advancements of leveraging large language model (LLMs)
for personalized tutoring and education  \cite{Chen2023EmpoweringPT,Kasneci2023-chatgpt,empowering-personalized-learning}. 
Despite efforts to develop technologies aimed at improving educators' oral presentation skills 
\cite{Asadi2017-intelliprompter,Okrasa2022-speech-delivery,Ochoa2022-multimodal}, 
a significant gap remains in utilizing technology 
to assist teachers to implement the scientifically-based teaching strategies 
in engaging students and enhancing learning outcomes. 

In educational settings, a teacher's emotional expressiveness has been linked to student motivation, comprehension, and retention 
of information \cite{FALCON2023101750}. Hence, accurately detecting and responding to these vocal sentiments 
can be pivotal in creating a more engaging 
lecturing environment. The integration of voice sentiment analysis in educational technology builds upon foundational research in 
affective computing and speech processing \cite{ARYA2021100399,gowda2017-affective,Lee2014}. 
Previous studies have demonstrated that emotional cues in speech, such as tone, pitch, and rhythm, can significantly 
impact communication efficacy and audience engagement \cite{Huang2021}. Speech emotion recognition (SER) 
is a task to determine the speaker's emotional state, such as happiness, anger, sadness, or frustration, from speech 
patterns like tone, pitch, and rhythm \cite{GEORGE2024127015}. 
Deep learning approaches have become prominent solutions for SER in recent years. Various neural network architectures have 
been applied \cite{Liu2023-speech}. 
Feature extraction is a crucial step in SER. Common features used include Mel-frequency Cepstral Coefficients (MFCCs)
and spectral features \cite{Singh2023,AOUANI2020251}.

%
%
\section{Background}
\label{sec:background}

\begin{figure*}[t]
	\centering
	\includegraphics[width=.6\textwidth]{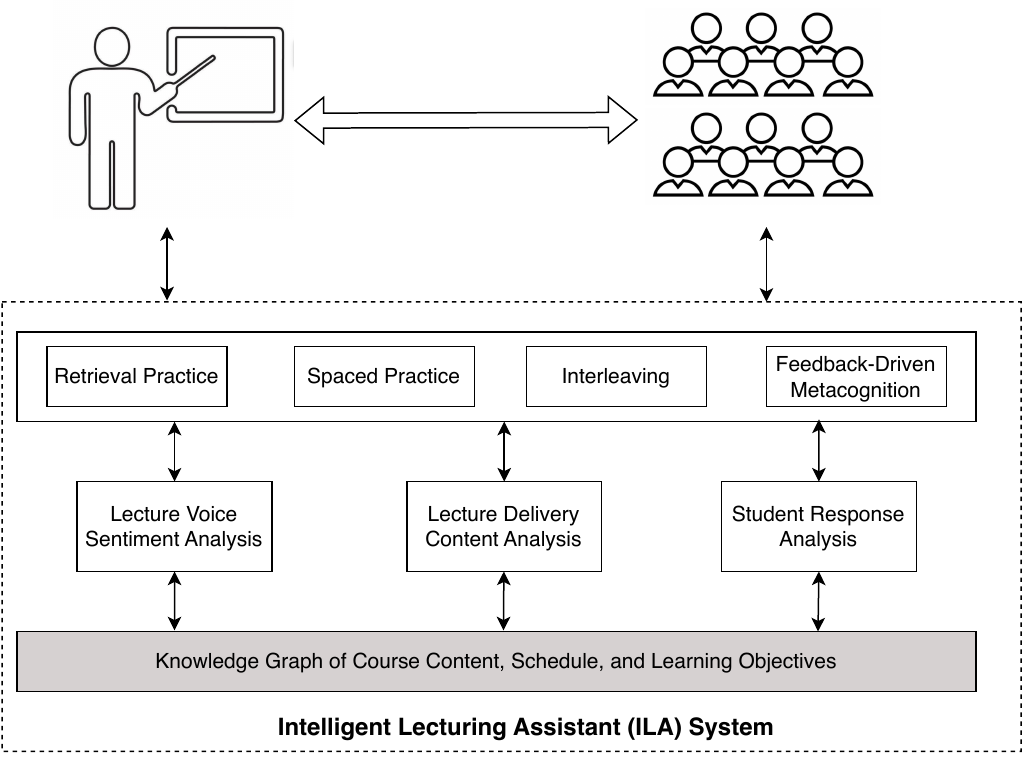}
	\caption{Overview of Intelligent Lecturing Assistant (ILA) System.
A knowledge graph that compiles course content, schedule, and learning objectives forms the system's foundation. 
The system constantly analyzes input from both the lecturer and students, including the lecturer's voice 
for sentiment, the actual content being delivered, and student responses through various means.
In the context of teaching and learning, the system applies the analysis
results to assist the teacher to implement the evidence-based learning strategies like retrieval practice, spaced practice, interleaving,
and feedback-driven metacognition.
 }
	\label{fig:aila_overview}
\end{figure*}

In this section, we provide an overview of the knowledge graph-supported intelligent lecturing assistant (ILA) system, 
with lecture sentiment analysis functioning as a critical component.
Figure \ref{fig:aila_overview} illustrates the overview of the system.
At its core lies a knowledge graph that compiles course content, schedule, and learning objectives. 
The system constantly analyzes input from both the lecturer and students, including the lecturer's voice 
for sentiment, the actual content being delivered, and student responses through various means.
In the context of teaching and learning, the system applies the analysis
results to assist the teacher to implement the scientifically-based 
teaching strategies like retrieval practice, spaced practice, interleaving,
and feedback-driven metacognition.
\emph{Knowledge graphs, machine 
learning, and speech recognition technologies play crucial roles in the development of such a system. }

Intelligent educational systems (IES) and intelligent tutoring systems (ITS) have leveraged knowledge graphs to 
address several key aspects of learning, including exercise recommendation and selection \cite{li2023-kg-exercise},
student interactions diagnosis \cite{SU2022109547},
personalized learning guidance \cite{Lu2021}, 
multi-dimensional knowledge representation \cite{li2023-kg-exercise},
integration with other AI techniques \cite{li2023-kg-exercise,SU2022109547},
and automatic grading \cite{Lu2021}.
By leveraging the structured representation of knowledge graphs, 
these systems can better assess student knowledge, recommend appropriate learning materials, and guide the learning process.

%
%
\section{The Problem of Lecture Voice Sentiment Analysis}
\label{sec:problem_sentiment_analysis}

The problem of lecture voice sentiment analysis is to classify whether the lecture speech in a certain period of time as engaging or not engaging. 
Formally, Let $S$ represent a lecture speech with a duration of $T$ minutes. 
Let $y$ denote the label associated with speech $S$, where $y\in\{0, 1\}$. Here, $y=0$ 
indicates that the lecture is "engaging," and $y=1$ indicates that the lecture is "not engaging."
We want to train a model $M$ such as $M(S)=y\in\{0, 1\}$.

We train the model, $M$, using a set of 1-minute voice clips, considering shorter clips allowing for more focused
and relevant acoustic  feature extraction. Also, a data set consisting of all short 1-minute clips makes
the data consistent and easier to apply data augmentation techniques (e.g., adding noise, varying pitch) to reduce overfitting.
In particular, we divide $S$ into $T$ segments, each of which has 1-minute duration. Let
$S=\{s_1, s_2, ..., s_T\}$ denote these $T$ segments, where $s_i$ is the $i^{th}$ 1-minute segment of $S$. 
For each $s_i$, the model produces a binary classification as $\hat{y}_i = M(s_i)\in\{0, 1\}$. 
We compute the average classification score across all $T$ segments as:
\[
\text{Score}(S) = \frac{1}{T} \sum_{i=1}^T \hat{y}_i
\]
The final classification of the entire lecture speech $S$ is determined based on whether the average score exceeds a threshold of 0.5:
\[
\text{Classify}(S) = 
\begin{cases} 
\text{engaging} & \text{if } \text{Score}(S) > 0.5 \\
\text{not engaging} & \text{otherwise}
\end{cases}
\]

In next section, we describe the process and results of collecting and labeling 1-minute lecture voice clips used for training the model. 

%
%
\section{Building Data Sets for Lecture Voice Sentiment Analysis}
\label{sec:data_collection}

\subsection{Training Data}

Given the absence of an existing dataset specifically tailored for training a model for lecture voice sentiment 
analysis, we decided to create such a data set from scratch. Our objective was to compile a comprehensive 
set of 1-minute lecture voice clips, which would serve as the foundation for training and evaluating our model.

To compile the dataset, each researcher carefully watched publicly available lecture videos from several video
sharing or massive open online course sites. We use the following criteria to manually label a lecture video as 
\emph{`engaging'} or \emph{`boring'} (i.e., \emph{`not engaging'}). 
A lecture is labeled as \emph{`engaging'} if it posses the following characteristics:
\emph{clear enunciation, appropriate loudness, appropriate pacing, varying pitch, enthusiasm, pauses, stressing keywords, varying 
voice, appropriate rhythm, asking questions}, etc.. A lecture is labeled as \emph{`boring'} if it posses the following characteristics:
\emph{Mumbling, monotonous, low volume, slow pacing,  too fast, no breaks, lack of emphasis,  
dull voice, unenthusiastic tone}, etc..

We extracted the voice from a labeled lecture and write a Python program to segment it into 1-minute clips.
To ensure consistency and reliability in the labeling process, each researcher independently reviewed and labeled the 1-minute clips according to above criteria. 
Additionally, we also conducted cross validation whereby a subset of the labeled clips was reviewed by cross researchers 
to verify the accuracy and reliability of the labels.

In order to enhance the dataset and introduce variations, we also recorded our own voice clips. 
These clips were deliberately recorded to produce examples of both engaging and boring lectures. 
This supplemental data enriched the dataset by providing a broader range of speaking styles and engagement levels, 
thereby enhancing the model's ability to generalize across contexts.

\begin{figure}[!ht]
	\centering
	\includegraphics[width=.49\textwidth]{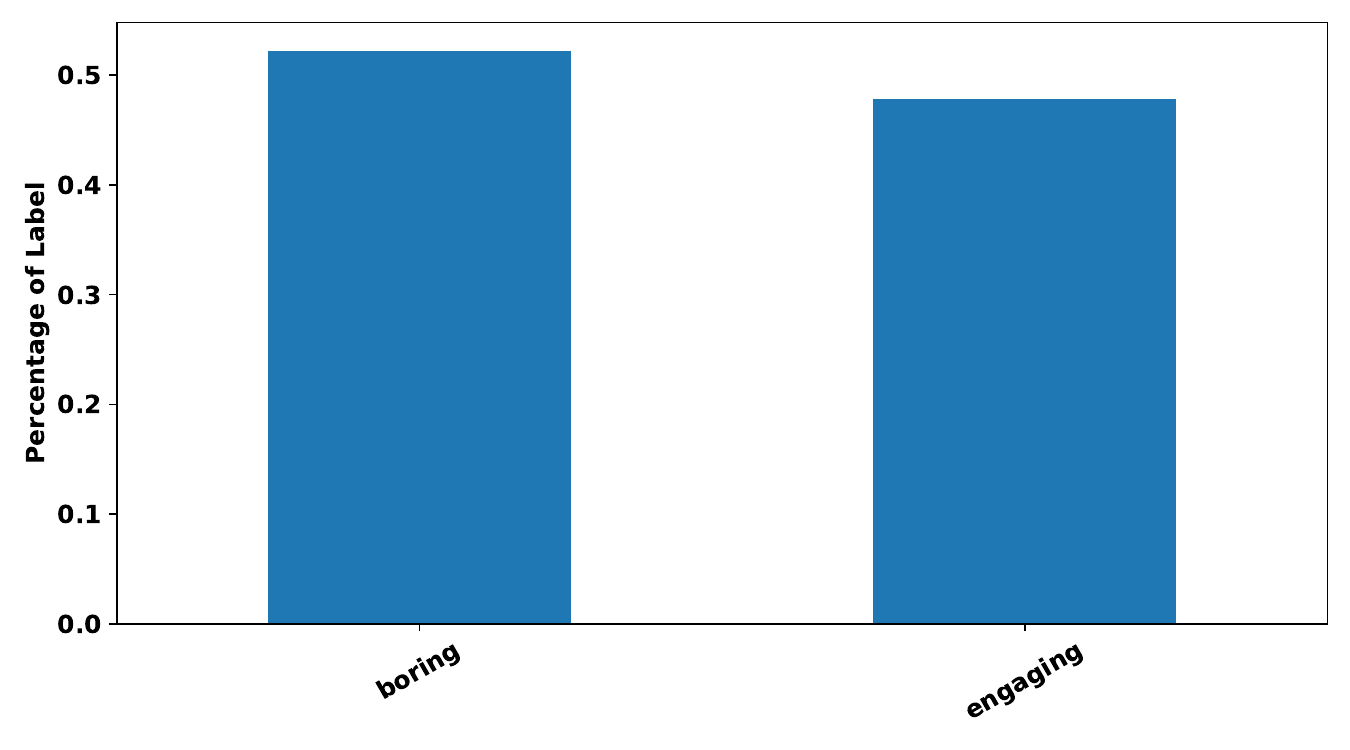}
	\caption{Label Distribution in the Data Set for Training}
	\label{fig:engaging-boring-distribution}
\end{figure}

Finally, we collected total 3,025 lecture voice clips. Table \ref{tab:clips-statistics} shows the numbers of 
clips collected by each researcher. Figure \ref{fig:engaging-boring-distribution} illustrates the distribution of the
clips labeled as \texttt{`engaging'} and \texttt{`boring'} by the researchers. It should be noted that the target 
labels are evenly distributed in the collected data set for the binary classification problem. 

\subsection{Independent Validation Data}

\begin{figure}[!ht]
	\centering
	\includegraphics[width=.49\textwidth]{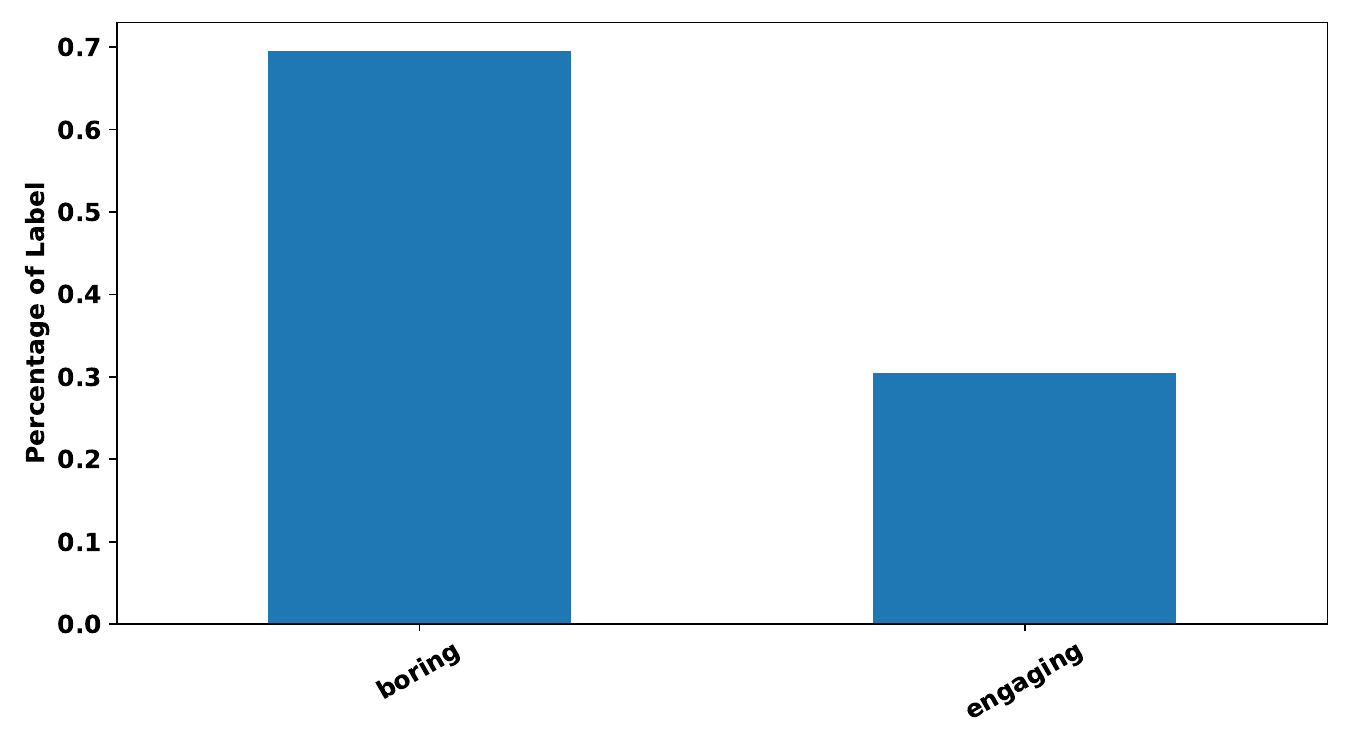}
	\caption{Label Distribution in the Independent Validation Data Set}
	\label{fig:test-data-distribution}
\end{figure}

The set of 1-minute clips used for training may contain multiple clips from the same speaker. This can potentially lead the model 
to learn and memorize specific voice characteristics instead of generalizable sentiment features. 
Consequently, the evaluation results may not accurate reflect the performance of the model.
To mitigate this issue, we collected a separate set of validation data that is independent of the training set, 
ensuring that the validation data does not include any speakers present in the training set. 
We will use the independent validation set to evaluate model performance for feature selection and hyperparameter tuning.
The independent validation set contains total 804 voice clips. Figure \ref{fig:test-data-distribution} illustrates the 
binary label distribution in the independent validation set.

\begin{table}[!ht]
    \centering
    \begin{tabular}{|c|c|c|c|}
    \hline
           \textbf{Researcher} & \textbf{Collection}  & \textbf{\# Engaging Clips} & \textbf{\# Boring Clips} \\
    
             & \textbf{Date}  &  &  \\
    \hline
		No.1 & 06/17/2024- & 644 & 363 \\
    
             & 07/12/2024 &  &  \\
    \hline
    		No.2 & 06/17/2024- & 201 & 233 \\
                & 07/12/2024 &  &  \\
    \hline
		No.3 & 06/17/2024- & 185 & 534 \\
            & 07/12/2024 &  &  \\
    \hline
		No.4 & 06/17/2024- & 200 & 200 \\
            & 07/12/2024 &  &  \\
    \hline
		No.5 & 06/17/2024- & 217 & 248 \\
            & 07/12/2024 &  &  \\
    \hline
    \end{tabular}
    \caption{Data Counts Collected by Researchers. We had in total 5 researchers (No.1 - 5) 
    who contributed to the data collection process.}
    \label{tab:clips-statistics}
\end{table}

%
%
\section{Extracting Features from Raw Voice Signals}
\label{sec:feature_extraction}

Raw audio signals represent sound waves captured over time, typically in the form of a one-dimensional array of amplitude values. 
Digitized audio signals are characterized by amplitude, duration, and sample rate that is the number of samples taken per second. 

\begin{figure}[!ht]
	\centering
	\includegraphics[width=.49\textwidth]{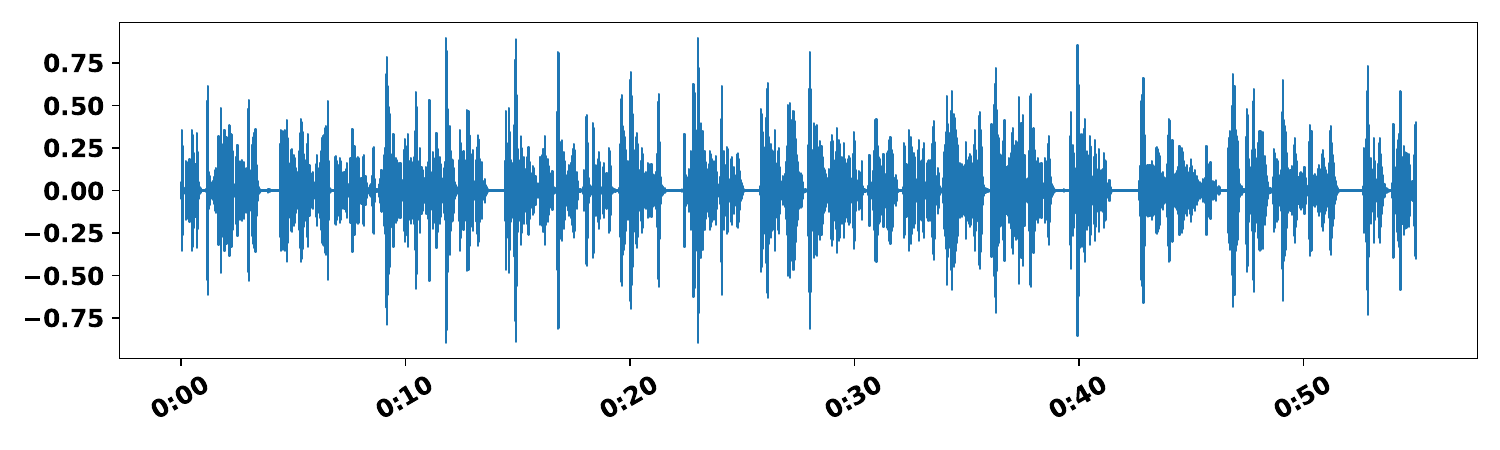}
	\caption{The Waveform Representation of an Example Lecture Voice Clip}
	\label{fig:example_waveform}
\end{figure}

Raw audio signals can be visualized using waveforms, which are graphical representations of the amplitude values over time.
Figure \ref{fig:example_waveform} shows the waveform of an example lecture voice clip.
While waveforms are useful for visual inspection, they do not directly reveal the frequency content or other more complex features 
such as pitch, loudness, or timbre, which are crucial for tasks like sentiment analysis.

To address these limitations, it is essential to extract meaningful features from the raw audio signals. 
For this purpose, we applied the \texttt{librosa} \cite{Mcfee2024-librosa} Python package to extract
a comprehensive set of features from the lecture voice clips, capturing various aspects of the 
audio signal that are relevant to sentiment analysis:

\begin{itemize}
\item Zero Crossing Rate (ZCR) \cite{Mcfee2024-librosa}: A measure of the rate at which the signal changes sign during the duration of a particular frame. 
High ZCR values typically indicate more noise or rapid changes in the audio signal, which can be correlated with certain speech characteristics.

\item Chroma STFT (Short-Time Fourier Transform) \cite{Ellis2007-chroma}: refers to the chroma feature representation derived from the short-time 
Fourier transform of an audio signal. Chroma features, or chromagrams, represent the energy distribution among the twelve 
different pitch classes (C, C\#, D, ..., B) of the musical octave. 
This feature captures harmonic content and can be useful in distinguishing different tones and pitches used by 
the speaker, which are often indicative of engagement levels.

\item Mel Spectrogram \cite{Rabiner2010-speech}: A representation of the power spectrum of a sound signal, where the frequencies are converted to the Mel scale. 
The Mel scale is designed to mimic the human ear's perception of sound, where each Mel unit corresponds to a perceived equal step in pitch. 
This feature provides a time-frequency representation that emphasizes perceptually relevant aspects of the speech signal, aiding in 
the analysis of its emotional and engaging content.

\item Mel Frequency Cepstral Coefficients (MFCC) \cite{hossan2010-mfcc}: A cepstral representation where the frequency bands are distributed according to 
the Mel scale. They are particularly effective in capturing the timbral characteristics of speech, which can be crucial for sentiment analysis.

\item Root-Mean-Square (RMS) Value \cite{Mcfee2024-librosa}: Computed either from the audio samples or from a spectrogram, 
provides a measure of the signal's power. 
RMS is indicative of the loudness of the speech, which can be a significant factor in determining the engagement level of the lecture.

\item Chroma CQT (Constant-Q Transform) \cite{Mueller2011-chroma}: Captures the pitch class energy distribution using a logarithmically spaced 
frequency axis, providing robustness to variations in the speech signal.

\item Chroma CENS (Chroma Energy Normalized Statistics) \cite{Mueller2011-chroma}: A variant that emphasizes long-term tonal content and is 
robust to variations in dynamics and articulation.

\item Chroma VQT (Variable-Q Transform) \cite{Mueller2011-chroma}: Similar to CQT but with variable-Q factor, providing flexibility in analyzing different frequency ranges with varying resolutions.

\item Spectral Centroid \cite{Klapuri2007-signal}: Represents the center of gravity of the spectrum, indicating where the majority of the signal's energy is located.

\item Spectral Bandwidth \cite{Klapuri2007-signal}: Measures the width of the spectrum, providing information about the range of frequencies present in the speech signal.

\item Spectral Contrast \cite{music-type-classification-spectral}: Captures the difference in amplitude between peaks and valleys in the spectrum, reflecting the harmonic structure of the speech.

\item Spectral Flatness \cite{Dubnov2004-generalization}: Measures the flatness of the spectrum, distinguishing between tonal and noisy signals.

\item Spectral Rolloff \cite{Mcfee2024-librosa}: The frequency below which a specified percentage of the total spectral energy lies. It is used to approximate the maximum or minimum frequency content of the signal.

\end{itemize}

The raw audio signal is typically divided into small, overlapping frames to analyze the time-varying spectral properties. 
For a feature that is computed across frames, we compute the mean across all frames to 
obtain a more stable representation for the entire audio clip. Table \ref{tab:feature_dimension} shows the feature names and their final dimensions.

\begin{table*}[!ht]
    \centering
    \begin{tabular}{|l|l|l|}
    \hline
           \textbf{Feature Name} & \textbf{Feature Description}  & \textbf{Dimension} \\
    \hline
		\texttt{zcrate\_mean} & Zero Crossing Rate (ZCR) & 1 \\
    \hline
    		\texttt{chroma\_stft\_mean} & Chroma STFT (Short-Time Fourier Transform)  & 12 \\
    \hline
		\texttt{melspectrogram\_mean} & Mel Spectrogram & 128 \\
    \hline
		\texttt{mfcc\_mean} & Mel Frequency Cepstral Coefficients (MFCC) & 20 \\
    \hline
		\texttt{rms\_mean} & Root-Mean-Square (RMS) & 1 \\
    \hline
    		\texttt{chroma\_cqt\_mean} & Chroma CQT (Constant-Q Transform) & 12 \\
    \hline
    		\texttt{chroma\_cens\_mean} & Chroma CENS (Chroma Energy Normalized Statistics)  & 12 \\
    \hline
		\texttt{chroma\_vqt\_mean} & Chroma VQT (Variable-Q Transform) & 12 \\
    \hline
		\texttt{spcent\_mean} & Spectral Centroid & 1 \\
    \hline
		\texttt{spband\_mean} & Spectral Bandwidth & 1 \\
    \hline
    		\texttt{spcontrast\_mean} & Spectral Contrast & 7 \\
    \hline
		\texttt{spflat\_mean} & Spectral Flatness & 1 \\
    \hline
		\texttt{sprolloff\_mean} & Spectral Rolloff & 1 \\
    \hline
    \end{tabular}
    \caption{Features and Dimensions; the Total Dimension is 209}
    \label{tab:feature_dimension}
\end{table*}

\begin{figure*}[!ht]
	\centering
	\includegraphics[width=.7\textwidth]{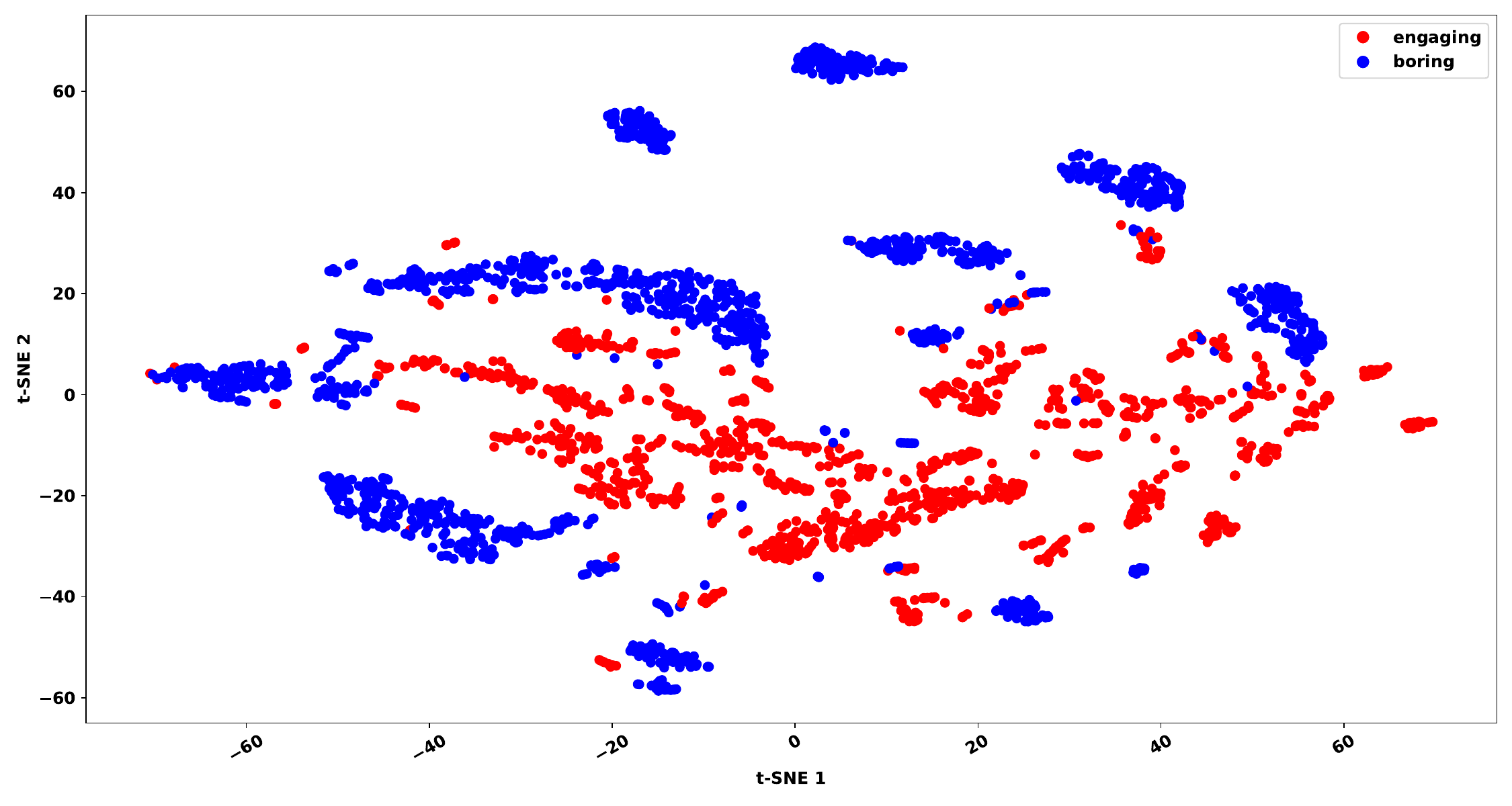}
	\caption{The Visualization of the Clips Based on Their MFCC Features}
	\label{fig:mfcc_plot}
\end{figure*}

Prior to building models, it is critical to visualize the clips based on their features. 
The primary goal is to develop a robust and accurate lecture voice sentiment analysis model by 
understanding data distribution. 
We chose MFCC features which are widely used in audio processing for the illustration purpose.
Figure \ref{fig:mfcc_plot} shows the t-SNE plot of the MFCC features. In the reduced-dimensional space,
the t-SNE plot reveals distinct red and blue clusters of engaging and boring clips. 
The points tend to form separate clusters, showing a clear distinction between engaging and boring clips.
Some overlap between the clusters may occur, highlighting the inherent variability in lecture delivery and the challenges 
in classifying borderline cases.

%
%
\section{Building Models for Lecture Voice Sentiment Analysis}
\label{sec:model_building}

Given the 13 extracted features, we focus on building and evaluating various classification models to identify 
the most effective approach for lecture voice sentiment analysis. Our strategy involves experimenting with traditional machine 
learning models as well as  
deep neural networks. For traditional model, we select three widely-recognized models, namely, \emph{logistic regression, random forest}, 
and \emph{XGBoost}.
For deep neural networks, we explore a \emph{fully connected network} and a \emph{convolutional neural network (CNN)}. 

\textbf{Feature Combinations:} To identify the best feature set for classification, we evaluate all possible combinations of the 13 features. 
This exhaustive approach ensures that we explore the potential synergies and interactions between different features, 
allowing us to select the most informative subset. Given that there are 13 features, the total number of possible subsets is 
$2^{13}=8192$ including the empty set. For each subset (excluding the empty set), we evaluate a model on the independent validation 
set to find the best hyper-parameters. This approach also provides valuable insights into which features are most critical for 
distinguishing between engaging and non-engaging lecture clips, informing future research and model development efforts.

\textbf{Model Selection:} For each of the traditional models, we systematically optimize key hyper-parameters to enhance their performance. 
This involves tuning parameters such as the regularization strength in Logistic Regression, the number of trees and minimum samples for 
splitting in Random Forest, and the learning rate and number of trees in XGBoost.

For deep neural networks, we switch between a fully connected neural network and a 
convolutional neural network depending the dimension of the input features. 
If the input dimension is smaller than 20, we build a fully connected neural networks where multiple dense layers are stacked to create a deep network. Each layer captures different levels of abstraction, allowing the model to learn complex representations from the input features.
If the input dimension is at least 20, we experiment with a convolution neural network whose architecture can capture local patterns in the feature 
space, which might be crucial for distinguishing between engaging and non-engaging lecture clips.
We selected 20 as the minimum number of dimensions for switching between a fully connected
neural network and a convolutional neural network because of empirical considerations and
avoiding overfitting.

Table \ref{tab:model_summary} show the summary of the models we explored in the study.

\begin{table*}[!ht]
    \centering
    \begin{tabular}{|l|l|l|}
    \hline
           \textbf{Model} & \textbf{Type}  & \textbf{Specification} \\
    \hline
		\texttt{logistic regression (lr)} & traditional & \emph{software}: \texttt{scikit-learn} \\
		& & \emph{parameters}: \texttt{C=[0.01,0.1,1,10,100],} \\
		& & \texttt{solver=`lbfgs', max\_iter=500,} \\
		& & \texttt{others: default.} \\
    \hline
    		\texttt{random forest (rf)} & traditional  & \emph{software}: \texttt{scikit-learn} \\
		& & \emph{parameters}: \texttt{n\_estimators=[200,500],} \\
		& & \texttt{min\_samples\_split=[2,10],} \\
		& & \texttt{others: default.} \\
    \hline
		\texttt{XGBoost (xgboost)} & traditional & \emph{software}: \texttt{scikit-learn, xgboost}  \\
		& & \emph{parameters}: \texttt{n\_estimators=[200,500],} \\
		& & \texttt{learning\_rate=[0.01,0.2],} \\
		& & \texttt{others: default.} \\
    \hline
		\texttt{fully connected network (denseNet) } & neural network & \emph{software}: \texttt{tensorflow, keras}  \\
		& & \emph{network structure}: \texttt{Dense(32,`relu'),} \\
		& & \texttt{Dense(16,activation=`relu'),} \\ 
		& & \texttt{Dense(1,activation=`sigmoid').} \\
    \hline
		\texttt{convolutional network (cnn)} & neural network & \emph{software}: \texttt{tensorflow, keras} \\
		& & \emph{network structure}: \texttt{Conv1D(32,3,`relu'),} \\
		& & \texttt{MaxPooling1D(2),} \\
		& & \texttt{Conv1D(64,3,`relu'),} \\
		& & \texttt{MaxPooling1D(2),} \\
		& & \texttt{Dense(64,activation=`relu'),} \\
		& & \texttt{Dropout(0.5),} \\ 
		& & \texttt{Dense(1,activation=`sigmoid').} \\
		
    \hline
    \end{tabular}
    \caption{The Summary of the Traditional Models and the Neural Networks Evaluated for Lecture Voice Sentiment Analysis}
    \label{tab:model_summary}
\end{table*}

%
%
\section{Experimental Results}
\label{sec:evaluation}

We conduct a set of comprehensive experiments. 
For each model, we evaluate it on a combination of features. If it is a type of traditional model, we use 
the independent validation set to select
the best parameters. We use \emph{Accuracy, Precision, Recall}, and \emph{F1-Measure} for evaluation metrics. 
For the lecture voice sentiment analysis problem, we are more interested in correctly 
identifying a boring lecture for intervention. We set the label \texttt{boring=1} as the positive case.
The metrics with respect to \texttt{boring=1} are defined as follows:
\[
\text{Accuracy} = \frac{TP+TN}{TP + FP+TN+FN}
\]

\[
\text{Precision} = \frac{TP}{TP + FP}
\]

\[
\text{Recall} = \frac{TP}{TP + FN}
\]

\[
\text{F1-Measure} = \frac{1}{\frac{1}{\text{Precision}} + \frac{1}{\text{Recall}}}
\]

In these formulas:
\begin{itemize}
\item $TP$ stands for true positives (correctly classified as boring),

\item $TN$ stands for true negatives (correctly classified as engaging),

\item $FP$ stands for false positives (incorrectly classified as boring),

\item $FN$ stands for false negatives (incorrectly classified as engaging).
\end{itemize}

The metrics with respect to \texttt{engaging=0} can be defined in the same way.

Table \ref{tab:evaluation_results} presents the best evaluation results for each model type specified in 
Table \ref{tab:model_summary}. 
Specifically, Table \ref{tab:evaluation_results} lists the parameters of the optimal model and the feature combinations 
that yielded the highest F1-measure score for the \texttt{boring} label. Additionally, the table includes accuracy 
and other metrics for the \texttt{engaging} label for reference.
In Table \ref{tab:evaluation_results}, the metric \textbf{Precision\_1} refers to the
metric calculated for label 1 (\texttt{boring}). This interpretation applies similarly to other metrics with suffixes.

\begin{table*}[!ht]
    \centering
    \begin{tabular}{|l|l|l|l|l|l|l|l|l|l|}
    \hline
	\textbf{Model} & \textbf{Parameters} & \textbf{Features}  & \textbf{Accuracy} & \textbf{Precision\_1\textsuperscript{*}}  & \textbf{Recall\_1} & \textbf{F1\_1} & \textbf{Precision\_0}  & \textbf{Recall\_0} & \textbf{F1\_0} \\
    \hline
    	\texttt{lr} & C=10 & zcrate\_mean & \textbf{0.87} & \textbf{0.94} & \textbf{0.86} & \textbf{0.90} & \textbf{0.74} & \textbf{0.88} & \textbf{0.80} \\
	&  & chroma\_vqt\_mean &  & & & &  & & \\
	&  & spcent\_mean & & & & &  & & \\
	&  & spband\_mean & & & & &  & & \\
\hline
	\texttt{rf} & n\_estimators=200 & rms\_mean & 0.67& 0.84 & 0.65 & 0.73 & 0.47 & 0.72 & 0.57 \\
	& min\_samples\_split=2 & spband\_mean & & & & &  & & \\
\hline
	\texttt{xgboost} & n\_estimators=200 & spband\_mean & 0.73 & 0.87 & 0.71 & 0.79 & 0.54 & 0.76 & 0.63 \\
	& learning\_rate=0.01 &  & & & & & &  & \\
\hline
	\texttt{denseNet} & See Table \ref{tab:model_summary} & zcrate\_mean & 0.81& 0.91 & 0.80 & 0.86 & 0.65 & 0.83 & 0.73 \\
	&  & spcent\_mean &  & & & &  & & \\
	&  & spband\_mean & & & & &  & & \\
	&  & spflat\_mean & & & & &  & & \\

\hline
	\texttt{cnn} & See Table \ref{tab:model_summary}  & rms\_mean & 0.75& 0.85 & 0.79 & 0.82 & 0.59 & 0.67 & 0.63 \\
	&  & chroma\_cens\_mean & & & & & &  & \\
	&  & chroma\_vqt\_mean &  & & & &  & & \\
	&  & spband\_mean & & & & &  & & \\
	&  & spflat\_mean & & & & &  & & \\
	&  & sprolloff\_mean & & & & &  & & \\
\hline    
\end{tabular}
    \caption{The Best Evaluation Results of Each Type of Model and Features. \\ 
    	\textsuperscript{Note *: 
    	The metric \textbf{Precision\_1} refers to the metric calculated for label 1 (boring). The same interpretation applies to other metrics with suffixes.}}
    \label{tab:evaluation_results}
\end{table*}

The results indicate that the logistic regression model with a regularization parameter of $C=10$ outperformed others 
across all evaluation metrics. The optimal feature set for this model includes the following four features: \texttt{zcrate\_mean}, 
\texttt{chroma\_vqt\_mean}, \texttt{spcent\_mean}, and \texttt{spband\_mean}. Among these, \texttt{zcrate\_mean}, \texttt{spcent\_mean}, 
and \texttt{spband\_mean} are single-value features, whereas \texttt{chroma\_vqt\_mean} is a vector with 12 dimensions. Analyzing the 
best-performing models across different metrics reveals that \texttt{spband\_mean} consistently appears in all top feature sets. 
Other features frequently found in high-performing models include \texttt{spcent\_mean}, \texttt{chroma\_vqt\_mean}, \texttt{zcrate\_mean}, 
and \texttt{rms\_mean}. These results suggest that spectral features are particularly effective for distinguishing between engaging 
and non-engaging lecture voices.

\textbf{Time Complexity:} On an Apple M2 MacBook Pro with 16 GB of memory, it took over 40 hours to select and evaluate the 
traditional models and over 30 hours to evaluate the neural networks. The model selection and evaluation 
processes were conducted across all 8,191 feature combinations.

\begin{figure*}[!ht]
	\centering
	\includegraphics[width=.7\textwidth]{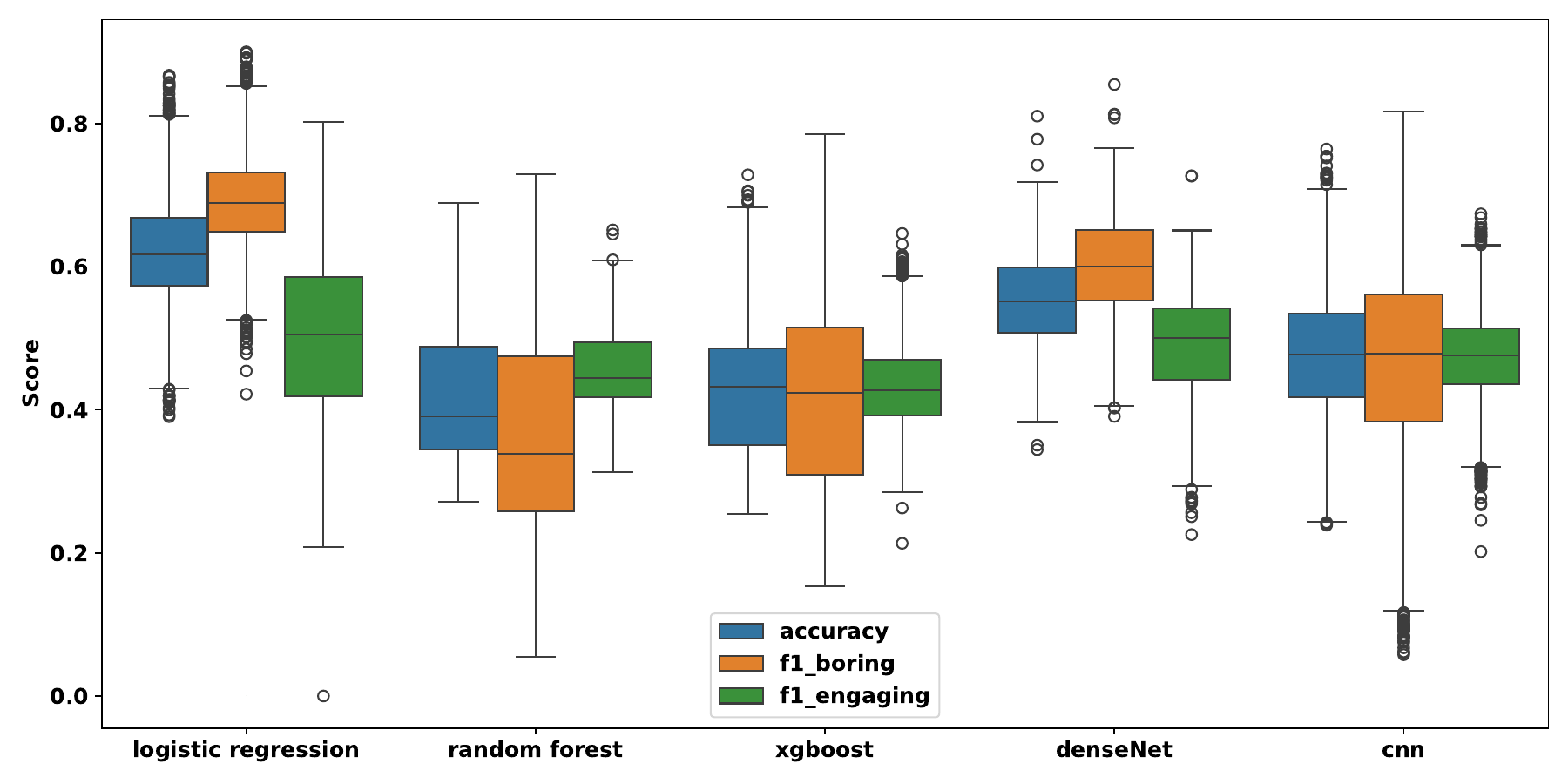}
	\caption{Accuracy and F1-Measure (Boring and Engaging) Scores over All Feature Combinations by Model.
 Each box shows the score distribution of the metric evaluated by the best model 
on all combinations of features.}
	\label{fig:accuracy_f1_plots}
\end{figure*}

To further investigate the performance of each model type, we plotted all the accuracy and F1-measure scores. 
Figure \ref{fig:accuracy_f1_plots} shows the box plots of accuracy and f1-measure scores. 
Each box shows the score distribution of the metric evaluated by the best model 
on all combinations of features. 
The plots confirm that the logistic regression model consistently 
outperformed other models across the various feature combinations.

%
%
\section{Discussion}
\label{sec:discussion}

The analysis of lecture voice sentiment analysis presents promising outcomes; however, the current study is 
limited by its focus on a binary classification of engagement (engaging vs. non-engaging), which may not fully capture 
all aspects of the lecturer delivery. Additionally, the model development and evaluation are constrained by the use of 
a dataset exclusively collected by the researchers of this study, which may limit the generalizability of the findings. 
Future research could benefit from expanding the dataset and incorporating more complex features to 
enhance the model's accuracy and robustness.

\emph{Ethical Considerations:} The analysis of lecture voice sentiment raises several important ethical 
issues that must be carefully considered. Firstly, the datasets could introduce biases inherent to 
the specific context in which the data was gathered. Such biases might limit the generalizability of 
the findings to different educational environments, demographics, or cultural contexts. 
Future research should strive to mitigate these concerns.
Secondly, the collection and analysis of lecture voice data involve sensitive information that 
could potentially be misused if not properly safeguarded. Ensuring the privacy is paramount.
Finally, the introduction of AI into educational contexts has the potential to exacerbate existing inequalities. 
It is crucial that AI systems are designed with inclusivity in mind, ensuring that they are fair and equitable 
across diverse teaching styles and populations.

%
%
\section{Conclusion}
\label{sec:conclusion}

The study presented in this paper explores the development of an intelligent lecturing assistant (ILA) system that utilizes a knowledge graph to 
represent course content and pedagogical strategies. The system is designed to support instructors in enhancing student engagement in learning
by analyzing lecture voice sentiment. The paper focuses on the development of a model that can classify lecture voice as either 
engaging or non-engaging, and the results demonstrate promising performance with an F1-score  of 90\% for boring lectures on an 
independent set of test voice clips.

This research on lecture voice sentiment analysis lays the groundwork for developing additional components 
of the intelligent lecturing assistant (ILA) system.
An ILA system is a multifaceted platform that integrates knowledge representation, reasoning, speech 
recognition, machine learning, and intelligent intervention. 
The next phase of development involves incorporating content analysis and pedagogical 
principles into the model, which would enable the system to deliver relevant interventions 
for instructors. Moreover, the system could be further developed to analyze student responses and provide real-time feedback.

The overall AI-powered  system has the potential to enhance student engagement and learning outcomes. 
Future research will also address ethical considerations related to the use of AI in education. 
The research team aims to deploy the system in both experimental environments and real classrooms to evaluate
the effectiveness of the system on both instructors and students.

%
%
\appendix

For the sake of reproducibility, the source notebooks utilized for data analysis and model evaluation are 
accessible in the public GitHub repository: https://github.com/anyuanay/KG\_ILA .

The training data and the independent validation sets can be made available for download upon request.

%
%
\bibliographystyle{unsrt}  
\bibliography{./references/aila,./references/my-publications}

\end{document}